\documentclass[runningheads]{llncs}
\usepackage{graphicx}
\begin{document}
\title{ICOS Protein Expression Segmentation: Can Transformer Networks Give Better Results?}
%

%
\author{Vivek Kumar Singh\inst{1} \and
Paul O’Reilly \inst{3} \and Jacqueline James \inst{1, 2} \and Manuel Salto-Tellez \inst{1} \and 
Perry Maxwell\inst{1}}
\authorrunning{F. Author et al.}
%
\institute{Precision Medicine Centre of Excellence, The Patrick G Johnston Centre for Cancer Research, Queen’s University Belfast, Belfast BT9 7AE, UK \and
Northern Ireland Biobank, The Patrick G Johnston Centre for Cancer Research, Queen’s University Belfast, Belfast BT9 7AE, UK \and Sonrai Analytics Ltd., Lisburn Road, Belfast BT9 7BL, UK}
\maketitle             
%
\begin{abstract}
Biomarkers identify a patient's response to treatment. With the recent advances in artificial intelligence based on the Transformer networks, there is only limited research has been done to measure the performance on challenging histopathology images. In this paper, we investigate the efficacy of the numerous state-of-the-art Transformer networks for immune-checkpoint biomarker, Inducible T-cell COStimulator (ICOS) protein cell segmentation in colon cancer from immunohistochemistry (IHC) slides.
Extensive and comprehensive experimental results confirm that MiSSFormer achieved the highest Dice score of $74.85\%$ than the rest evaluated Transformer and Efficient U-Net methods.

\end{abstract}

\section{Introduction}
Inducible Co-Stimulator (ICOS) may be a biomarker of interest in checkpoint inhibitor therapy and as a means of assessing T-cell regulation as part of a complex process of adaptive immunity.  
Over the last decade, deep learning-based convolutional neural networks (CNNs) have achieved outstanding performance in multiple applications \cite{dong2021survey}.
Capturing long-term dependencies requires increasing the size of the convolution kernel, which can slow down the system while improving the representation of functionality. Recently, Transformer-based networks developed that use self-attention mechanisms to extract long-distance dependencies without slowing them down. This demonstrates significant benefits for natural image classification, segmentation, and recognition tasks. Due to the higher parameters and the need of larger datasets in training, the work done in the medical field is limited \cite{shamshad2022transformers}. 
Therefore, the main motivation for this study is to measure the performance of the multiple Transformer methods, which capture a wide range of contextual information of different shapes and sizes of ICOS cells. 
\begin{figure}[!h]
\centering
\includegraphics[width=13cm, height=2.7cm]{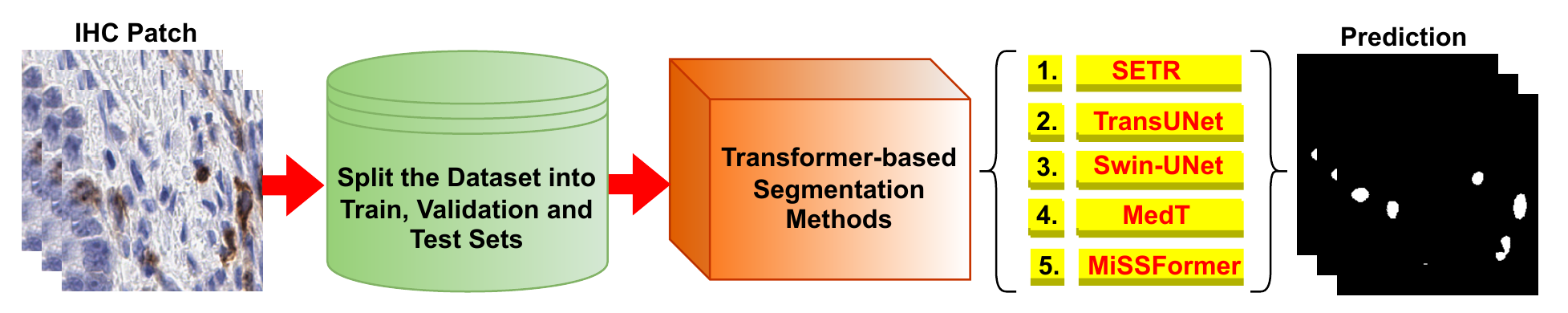}
\caption{Illustration of the proposed pipeline for ICOS cells segmentation.
\label{framework}}
\end{figure} 
Figure \ref{framework} shows the general pipeline of proposed framework.
To the best of our knowledge, we believe this is the first adaptation of analyzing the multiple Transformer network (i.e., SETR \cite{zheng2021rethinking}, TransUNet \cite{chen2021transunet}, Swin-UNet \cite{cao2021swin}, MedT \cite{valanarasu2021medical} and MiSSFormer \cite{huang2021missformer}) performance with self-attention mechanism in the application for ICOS cell segmentation from IHC images. 

\section{Material and Methods}
\textbf{Dataset:}
In this study, we used ICOS IHC data from our previous research \cite{sarker2021means}.
The entire dataset of 534 image patches is divided into training, validation, and test sets at 60\%, 10\%, and 30\%, respectively.

\textbf{Transformer-based segmentation methods:} we used the five  methods describe as follows.

\textbf{SETR} \cite{zheng2021rethinking} substitutes the convolutions in the encoder with a spatial resolution Transformer. The input is converted to a sequence of many  patches via learned patch embedding. It transformed through a global self-attention delivering an identifiable feature representation. 
    The decoder generates the binary mask. The source code is available at \url{https://github.com/920232796/SETR-pytorch}.
    
\textbf{TransUNet} \cite{chen2021transunet} incorporated the CNN-Transformer network. The CNN layers capture spatial information, and Transformer responsible for global feature. It has u-shape, where extracted self-attention features maps upsampled to be merged with varying CNN features skipped from the encoder. The implementation is available at \url{https://github.com/Beckschen/TransUNet}.

\textbf{Swin-UNet} \cite{cao2021swin} used the Swin Transformer in the encoder with shifted windows approach. The image is split into several patches and fed to the encoder extracting spatial and global features. The decoder uses a symmetric Swin Transformer consisting patch expanding layer to create the mask. The source code is available at \url{https://github.com/HuCaoFighting/Swin-Unet}.

\textbf{MedT} \cite{valanarasu2021medical} combines the extracted encoding layers local and global features. It applied a Local-Global training process for Transformers that utilize a narrower global path and a deep local path to works on the image patches. It leverages the axial attention technique that maintains the information passed to key, query, and value via a positional embedding. The implementation is available at \url{https://github.com/jeya-maria-jose/Medical-Transformer}.

\textbf{MISSFormer} \cite{huang2021missformer} used the enhanced Transformer block to obtain better features of ICOS cells. It includes the encoder, bridge, and decoder network with skip connection to narrow down the semantic gaps. The encoder layer extracts the relevant features of overlapped patches. The local and global features are combined via the bridge mechanism. The decoder network provide the segmentation map. The implementation is available at \url{https://github.com/ZhifangDeng/MISSFormer}.

\textbf{Implementation details:} We used the input size of $256 \times 256$ and applied the data augmentation with rotation, and horizontal flipping. An Adam optimizer was employed with a learning rate of 0.0002. We used the binary cross-entropy and dice losses. All the models has trained with a batch size of 4 with 50 epochs.

\section{Experimental Results}
Table \ref{Table1} shows the five state-of-the-art Transformer-based methods for ICOS cell segmentation. It includes the SETR \cite{zheng2021rethinking}, TransUNet \cite{chen2021transunet}, Swin-UNet \cite{cao2021swin}, MedT \cite{valanarasu2021medical} and MiSSFormer \cite{huang2021missformer}. Experimental results confirm that MiSSFormer has achieved the highest Dice and aggregated Jaccard index (AJI) scores of $74.85\%$ and $59.26\%$ respectively, than existing methods. This method allows capturing the long-range dependencies and local context of multi-scale ICOS cells features, improving segmentation results.
SETR has achieved the lowest Dice and AJI scores, which were $17\%$ lower against the MiSSFormer. We found that TransUNet, Swin-UNet, and MedT performed relatively well and were very similar in all the metrics.
We compared the Transformer methods to recent work \cite{sarker2021means} that achieved a $73.44\%$ Dice score, which was $1.5\%$ lower than MiSSFormer. Overall, Transformer-based method shows better segmentation results than CNN.
\begin{table}[!h]
\centering
\caption{Comparison of the latest Transformer-based segmentation models with ICOS.}
\label{Table1}
\scalebox{0.77}{
\begin{tabular}{|c|c|c|c|c|} 
\hline
Model      & Dice coefficient & AJI   & Sensitivity & Specificity  \\ 
\hline
SETR       & 57.90 & 41.95 & 64.47       & 99.03       \\ 
\hline
TransUNet  & 71.09 & 56.39 & 78.78       & 99.28      \\ 
\hline
Swin-UNet  & 71.52 & 55.61 & 77.92       & 99.42     \\ \hline
MedT       & 71.61 & 56.91 & 80.26       & 99.28      \\
\hline
\textbf{MiSSFormer} & \textbf{74.85} & \textbf{59.26} & \textbf{81.13}       & \textbf{99.49}   \\ 
\hline
\end{tabular}}
\end{table}

\begin{figure}
\centering
\includegraphics[width=13cm, height=3.5cm]{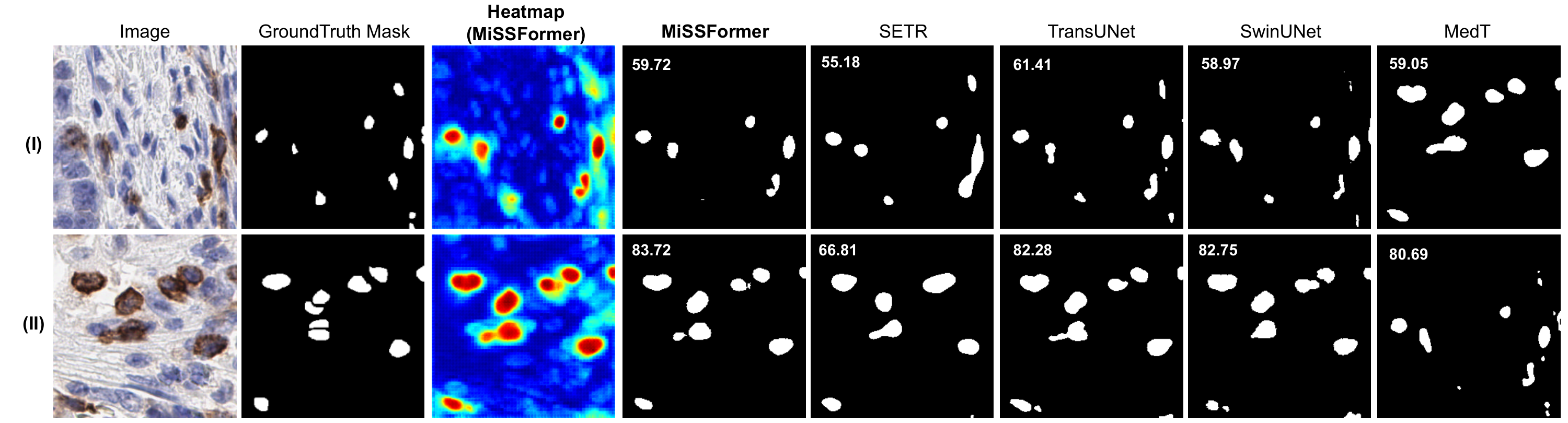}
\caption{Illustration of five Transformer methods for ICOS cells segmentation. 
\label{icos_example}}
\end{figure}

Figure \ref{icos_example} provide the qualitative comparison of Transformer-based methods. We provide the two examples has variability in cell structures.
The reported AJI score confirms that the MiSSFormer performed better than other methods and had fewer false positives. We also show the heatmap generated by the intermediate layer of the MiSSFormer. The higher red highlight network precisely captures the cells and ignores the background shown in blue.

\section{Conclusion}
This paper presents novel research to investigate Transformer-based networks for ICOS cell segmentation. We examined the effectiveness of five state-of-the-art segmentation methods.
Current studies comparing Transformer results to the recent CNN-based method show MiSSFomer attained improved segmentation results in all metrics. It can learn better feature representation than CNN. Future work aims to generalize the Transformer network to other protein cell datasets.

\bibliographystyle{splncs04}
\bibliography{bibliography} 

\end{document}